\documentclass[sigconf]{acmart}

\renewcommand\footnotetextcopyrightpermission[1]{} 

\setcopyright{none}
\copyrightyear{}
\acmYear{}
\acmDOI{}

\acmConference[Arxiv preprint]{Arxiv preprint}{2021}
\acmPrice{}
\acmISBN{}

\AtBeginDocument{%
  \providecommand\BibTeX{{%
    \normalfont B\kern-0.5em{\scshape i\kern-0.25em b}\kern-0.8em\TeX}}}

\begin{document}

\title{CHARTER: heatmap-based multi-type chart data extraction }
\author{Joseph Shtok*, Sivan Harary*, Ophir Azulai, Adi Raz Goldfarb, Assaf Arbelle, Leonid Karlinsky*}
\authornote{authors contributed equally to this research.}
\affiliation{%
  \institution{IBM Research AI, Haifa Research Lab}
  \streetaddress{Haifa Research Lab}
  \city{Haifa}
  \country{Israel}
}

\renewcommand{\shortauthors}{Shtok et.al.}

\begin{abstract}

The digital conversion of information stored in documents is a great source of knowledge. In contrast to the documents text, the conversion of the embedded documents graphics, such as charts and plots, has been much less explored. We present a method and a system for end-to-end conversion of document charts into machine readable tabular data format, which can be easily stored and analyzed in the digital domain. Our approach extracts and analyses charts along with their graphical elements and supporting structures such as legends, axes, titles, and captions. Our detection system is based on neural networks, trained solely on synthetic data, eliminating the limiting factor of data collection. As opposed to previous methods, which detect graphical elements using bounding-boxes, our networks feature auxiliary domain specific heatmaps prediction enabling the precise detection of pie charts, line and scatter plots which do not fit the rectangular bounding-box presumption. Qualitative and quantitative results show high robustness and precision, improving upon previous works on popular benchmarks.
\end{abstract}



\maketitle

\section{Introduction}
\begin{figure}
  \includegraphics[width=0.5\textwidth]{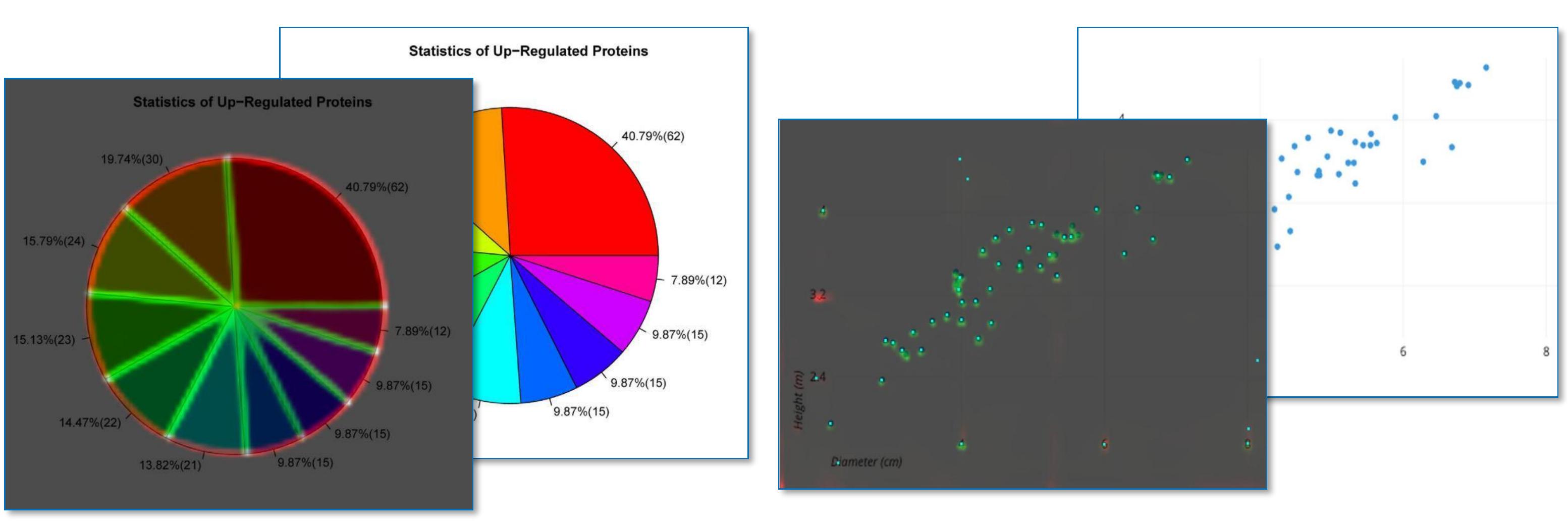}
  \caption{Heatmaps for non-rectangular graphical elements. }
  \label{fig:heatmaps_examples}
\end{figure}

Charts carry important part of the information content in many kinds of documents - financial reports, scholarly articles, presentations, to name a few. Recent interest in automatic document processing and conversion, such as in summarization and question answering tasks, has increased the importance of the extraction of underlying tabular data from chart images embedded in the converted documents. Chart analysis methods have evolved substantially in recent years from human-in-the-loop platforms relying on manual annotations
\cite{Shadish2009, Jung2017}, through early data extraction algorithms \cite{Al_Zaidy2015}, hybrid neural-algorithmic pipelines \cite{Poco2017,Dai2018}, to end-to-end processing by a neural network \cite{Liu2019c,Kafle2020,Zhou2021}. 

Commonly a two-stage approach is used, first detecting the chart regions in the documents, and then applying some data extraction process to the detected charts.
While the scope of the detection stage can be quite wide, including many types of charts \cite{Dai2018}, current tabular data extraction systems are mostly limited to the bar charts \cite{Al_Zaidy2015,Dai2018,Kafle2020,Zhou2021}, with few exceptions. One of the possible reasons is that standard object detectors, employed in recent works, better cope with (and enable easy inference from) objects like rectangular bars and text elements, less so with pie segments, while elements like line or area plots defy handling by box proposals. Additionally, matching of the chart legend to the corresponding chart elements, non-trivial due to small size of the color/texture chart component samples and the legend layout variability, remained out of scope for most of the recent works.

In this work we present a chart detection and analysis pipeline, addressing the mentioned limitations. We train a CNN object detector to produce predictions in the form of heatmaps, in addition to its standard bounding boxes proposals, thus enabling localization of graphical elements of arbitrary shape (as visualized in Figure ~\ref{fig:heatmaps_examples}). This allows us to better incorporate the geometrical domain knowledge of the chart structure into the detector during training, as well as extending the scope of chart conversion to bar, pie, line and scatter charts, all within the same (single) model. Beyond the improved detection of the graphical elements, we introduce algorithms for explicit label matching for bar and pie elements from colored legends, as well as from connector-based labels. 

Our main contributions are: (1) introducing heatmap prediction for general graphical chart elements; (2) presenting novel algorithms for data extraction from pie and line charts; (3) presenting quantitative results and an ablation study of the proposed methods, showing improvement over present benchmarks.  
\begin{figure*}
  \includegraphics[width=\textwidth]{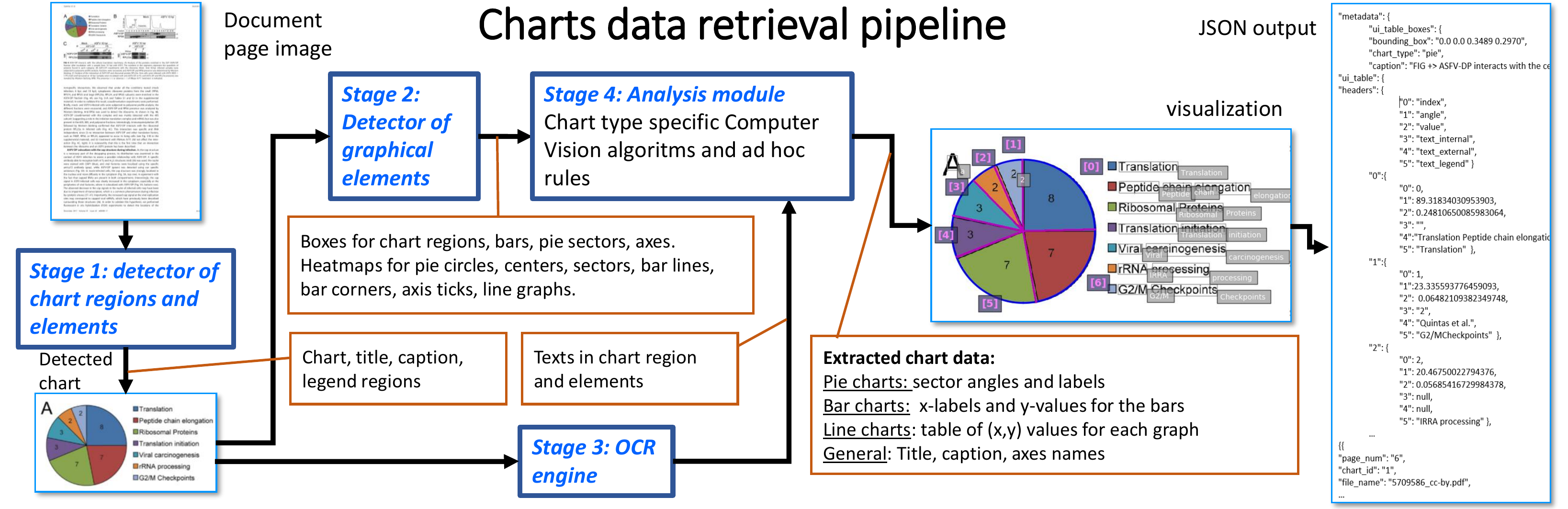}
  \caption{Our charts conversion pipeline. }
  \Description{The pipeline consists of three stages: detection of chart region, detection of graphical elements, OCR of the entire pare, and Analysis module integrating all the network outputs with Computer Vision algorithms and logic rules. }
  \label{fig:flowchart}
\end{figure*}
\section{Related work}
Early works addressing automatic charts classification and data extraction  \cite{Savva2011,Al_Zaidy2015}, used classical computer vision techniques, such as clustering image patches, Hough transform and OCR. 
More recently, \cite{Poco2017,Dai2018,Cliche2017} have presented hybrid neural-algorithmic pipelines, performing detection of the graphical objects with following extraction of numerical and textual information using OCR, Computer Vision techniques and rules; our approach belongs to this group of methods in terms of its general design. Other line of works \cite{Liu2019c,Kafle2020,Zhou2021} proposes an end-to-end analysis of the charts by a neural network. \cite{Zhou2021} develops an encoder-decoder architecture an attention mechanism for direct data extraction from bar charts by an RNN. \cite{Kafle2020} introduced a neural model for Question Answering about charts, and in particular they allow to extract tabular data by appropriate sets of questions.
In \cite{Liu2019c} a standard object detector is equipped with a relation network to address the connections between the different chart elements, 
this model is able to produce bar heights and angles of pie segments (for single pie chart), and to match them against the legend entries. However, this approach requires an individual network design for each type of charts, and the inference process involves affine transformations of the input image for spatial alignment of features.


We offer a lighter solution of detecting all the required kinds of graphical elements with a single network; however, the integration stage, applying domain knowledge to support a wide variety of real-world chart types, designs and issues, better fits the bill of a module built with logical rules and explicit algorithmics. It has enabled us to achieve a robust behaviour on a number of large real-world collections of documents.

\begin{figure*}
  \includegraphics[width=\textwidth]{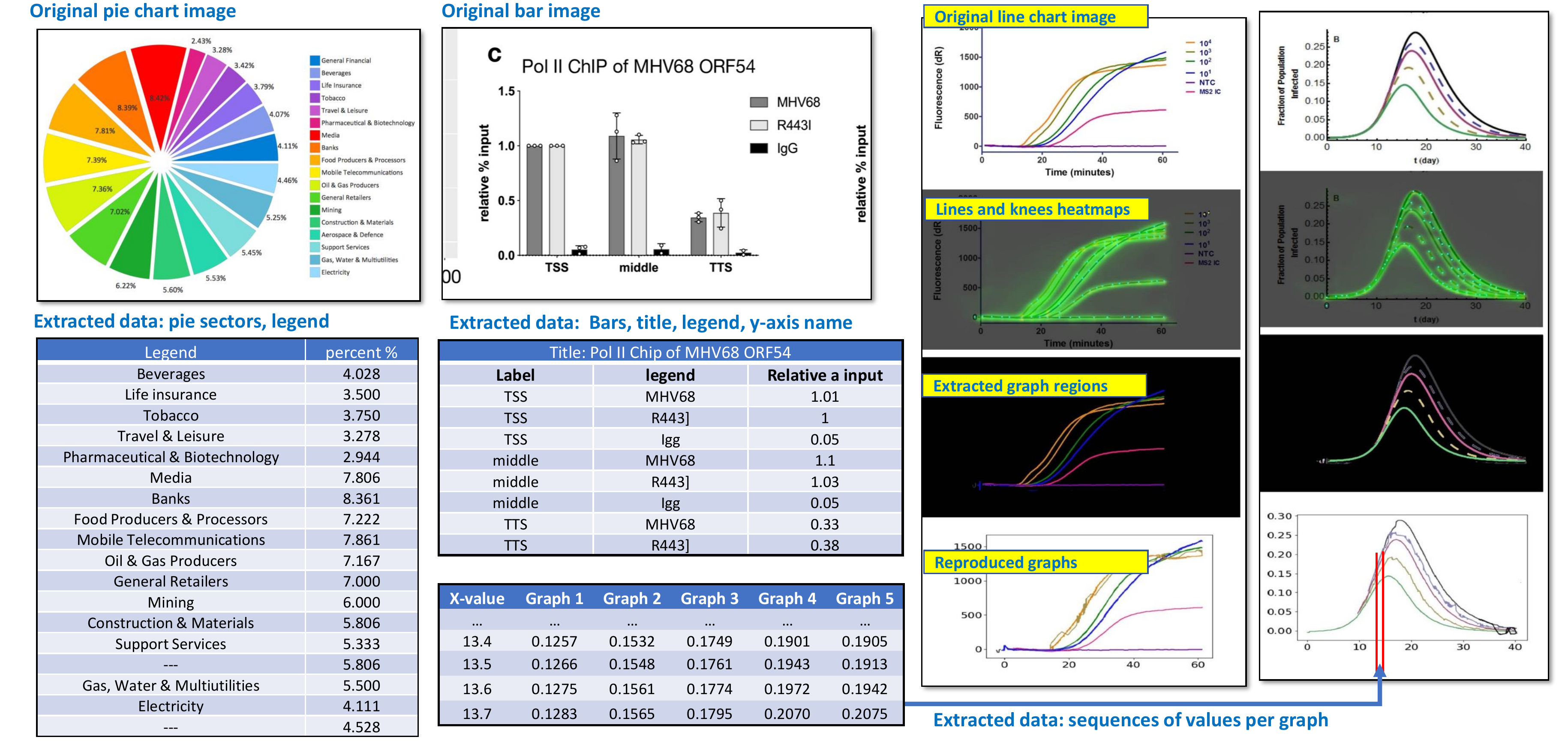}
  \caption{Examples of data extraction from different chart types. }
  \label{fig:chart_examples}
\end{figure*}

\vskip -1 cm
\section{Method}
In Stage 1 of our system an object detector is applied to complete document pages to retrieve regions of charts, title, legend, and caption 
\footnote{We have observed cases where these elements are shared by a group of charts in a figure, and are therefor not necessarily adjacent to any one of them; in this case, page-level logic is beneficial}. In stage 2, another object detector operates on the detected chart regions, producing both bounding box proposals and heatmaps for various chart components. It is trained to predict the important encapsulating regions (e.g. the outer and inner boxes around the chart) and some graphical elements (e.g. bars) using bounding boxes, while heatmaps are used to locate fiducials (e.g. axes ticks, pie junctions) and non-rectangular elements (e.g. circumference and radial lines in pie chart, line-plot graphs, etc.). In Stage 3, the chart image and text regions are processed by an OCR system for text extraction. Finally, in Stage 4, the Analysis module aggregates the data produced so far, applying computer vision algorithms and domain knowledge of different types of charts, to reproduce the  original source tabular data represented by the chart.
The system flowchart is presented in Figure~\ref{fig:flowchart}, while its details are provided in following subsections.

\subsection{Stage 1: Detection of chart regions}\label{sect:chart_detection} 
Our full-page chart regions detector is a Faster-RCNN model employing the FPN detection meta architecture and RN50 backbone implemented using the Detectron2 framework 
and trained on a real data.The object categories of the model are: bar chart, pie chart, line chart, scatter plot, legend, title, caption, and x- and y-labels. 

\subsection{Stage 2: Detection of graphical elements}
\textbf{Generation of synthetic charts data:}\label{sec:synth}
we extended the FiqureQA dataset \cite{Kahou2017} to generate the synthetic data for training the stage 2 detector. The chart generation variability was increased by: random background color, random border line color and style, random spaces between the bars, random bar bottom value, allowing uniform bar/slice color, allowing hidden axes etc. Moreover, we randomly add texture to the background, bars, and pie slices by pasting tiles from Describable Textures Dataset \cite{Cimpoi2014}. 

\textbf{Heatmap based detector:}
to analyze the content of the chart image and discover the graphical elements, we use the CenterNet model \cite{Zhou2019} modified to produce a number of heatmap types, in addition to the standard bounding box predictions. The model builds upon the stacked Hourglass-104 architecture and was trained only on synthetic chart images (Sec. \ref{sec:synth}).  
Categories supported by box predictions are vertical bar, horizontal bar, pie sector, and bar/pie chart regions (for cascade-style validation of Stage-1 predictions and refining the chart location). The categories supported by heatmaps are: four corners of the bar (in separate categories), x-ticks, y-ticks, center, circumference and radial lines of the pie chart, corners of the pie sectors, line-segment knee-points and lines of the line charts (which are in fact polygonal lines comprised of these segments), and dots of the scatter plots. The predictions made by Stage 2 model are later used in the Analysis module of Stage 4 for recovery of the chart tabular data.

\subsection{Stage 3: OCR engine}\label{sect:OCR}
Texts in charts may contain rotated X-axis labels and numbers with exponential notation. Our OCR pipeline is composed of the CRAFT text detection model \cite{Baek2019a}, producing text regions and angles, and the Clova AI recognition model \cite{Baek2019b} with added support for rotated text and numbers with exponents. 
The rotated text regions are handled appropriately before passing to the recognition model.
Since Clova AI was not originally designed to handle numbers with exponents, we first detect horizontally aligned text boxes where the text begins with the digits 10, and check if following digits are superscript.

\subsection{Stage 4: Analysis module}\label{sect:analysis}
The Analysis model builds upon recognized texts and graphical objects to discover the tabular data generating the chart. There is a dedicated logic for each type (bar, pie and line) of chart. Regretfully, we omit the details for line charts due to space limitation.

\textbf{Bar charts}\label{sec:bars}
We rely on Stage 2 detections of individual bars, while filtering them assuming equal bar width and a common y-level for one of the horizontal edges for all the bars. The bar heights are computed by first recovering the numerical y-axis (if available), via a Hough transform aligning the numerical OCR outputs on a vertical line, followed by interpolating the heights of the horizontal edges. Alternatively, we look for height values written inside or above the bars. Bar labels are read below the bars, or are retrieved from a legend by color matching; the treatment of textured legends is left for future work. The axis titles are detected by combining the detections of Stage-1 and Stage-2 models. Finally, the common chart elements, such as title and caption (found nearby by the Stage-1 detector) are added. The process was made robust by addressing the numerous special cases and issues encountered in the real world data, this robustness was then validated via results on the three datasets detailed in Section \ref{sect:experiments}.

\textbf{Pie charts: }\label{sect:analysis_pies}
 we use both box proposals and image-size heatmaps, produced by the CenterNet detector \cite{Zhou2019}, for the detection of pie circle center, radius and the radial lines. Initial attempt to use only the sectors box proposals have proven insufficiently robust (it was difficult to filter proposals by the predicted score, especially for narrow sectors), although the circles were well fitted. Instead, we generate separate heatmaps for a circle circumference, the center point, the radial lines, and the sector corners; then the final center and pie circle are computed by initiating the centers local maxima of the center point heatmap (multiple pies are allowed), voting to a range of radii with the circle points heatmap values (each pixel votes to its distance from the center), and update the center locations as centers-of-mass of pixels at predicted radius. Finally, we detect the radial lines (disc sectors) using the radial lines heatmap within the established circles.
Using the recovered pie geometry, the sector labels are detected, if a legend is present, by matching the legend colors to those of sectors. Otherwise, we look for lines starting inside the pie disc and leading out (connected components crossing the pie edge), and extract the label using the text piece on the other end as a seed. Finally, if no connectors are present, we match texts located within/outside of the corresponding sectors.

\textbf{Line charts: }
for line charts, we use heatmaps for the graph lines, and heatmaps for knee points in the piece-wise linear graphs the model was trained on. The individual lines are extracted by color-based clustering of the highlighted region
; non-continuous line types are supported by a stitching algorithm building the output line from available pieces. As color clustering is not always accurate, we use noise filtering and morphological operations to robustify the inference. We extract x- and y- numerical axes (in same way as for the bar charts) to reconstruct the  numeric data (an example is presented in Figure \ref{fig:chart_examples}) 
as (x,y) pairs. 

\section{Results}\label{sect:experiments}
In this section we provide the quantitative analysis of different modules and an ablation study for the heatmap-based data extraction. 

\subsection{Datasets for training and evaluation} For stage 1 detector, we have collected and manually annotated 743 document pages, of which 122 were crawled from the web, and 521 taken from 
PubMed COVID-19 related medical articles. They contain 1260 bar charts, 336 pie charts, 256 line charts, and 213 scatter plots. Stage 2 detector was trained on a synthetic dataset, generated with FigureQA code, as detailed in Section \ref{sec:synth}. It consists of 30k chart images for each type (pie, vertical bar, horizontal bar, line, and scatter charts). Test data for this stage comprises of 169 real-world pie and bar charts, also partly crawled from the web, and partly taken from the COVID-19 documents. For the evaluation of the data extraction performance we have used 70 bar charts from ICPR2020 Competition data \cite{ICPR2020} and manually annotated 30 pie charts from this corpus.


\subsection{Evaluation of Stages 1,2}
We have trained  the stage 1  detector in the k-fold fashion, using 5 random 90\%/10\% train/test splits of the 743 pages. The results, averaged over the 5 splits, are given in Table \ref{tab:res_chart_det}. As reflected by the AP values, the performance of the stage 1 detector is satisfactory for the rest of the system to rely on.

\begin{table}
  \caption{Detection of charts and related elements in documents}
  \label{tab:res_chart_det}
  \begin{tabular}{c|ccccc}
    \toprule
    Category & Bar &  Pie & Line & Scatter &  \\
    
    AP@0.5 & 98.0\% & 97.8\% &  84.4\% & 91.35\% &   \\
     \midrule
     Category & Legend & Caption & Title & X label & Y label \\
     
     AP@0.5 & 77.8\%& 91.4\% &  71.8\%  & 84.7\%  & 94.9\% \\
     
    \bottomrule
  \end{tabular}
  \caption*{Stage 1 detector performance, measured as Average Precision (AP) with IoU$=0.5$, averaged over 5 random train/test splits.}
\end{table}

Stage 2 detector performance on the 169 real-world charts is given in the Table \ref{tab:res_graphical_det}). It is compared to \cite{Liu2019c}, where a similar evaluation was conducted (on their \textit{Annotated} dataset).
\begin{table}
  \caption{Detection of graphical elements in chart
image}
  \vskip -2 mm
  \label{tab:res_graphical_det}
  \begin{tabular}{c|ccc}
    \toprule
    Category &Horizontal bar & Vertical bar & Pie sector\\
    \midrule
    \cite{Liu2019c} & --& 80.2\% & -- \\
    Ours  & 76.5\% & 90.5\% & 90.9\% \\
    \bottomrule
  \end{tabular}
  \caption*{Performance (in Average Precision (AP) with IoU$=0.5$} of the graphical elements detector (box categories) on the real data.
\end{table}

 \subsection{Stage 4: extraction of source tabular data}\label{sect:exp_extraction}
 We evaluate the quality of recovering the tabular data, following the experiment designs in \cite{Zhou2021} and \cite{Liu2019c}. The test is conducted on the real-world data of 70 bar charts and 30 pie charts from the ICPR2020 dataset, and is compared to similar Annotated datasets loc.cit. (30 charts in \cite{Zhou2021} and 10 bar charts + 10 pie charts in \cite{Liu2019c}).
 The accuracy of numeric information (bar heights and pie sectors angles) is measured as a portion of detected elements
 which numeric value is recovered within $\epsilon$ relative error:  $\left|(val_{GT}-val_{pred})/{val_{GT}}\right|\leq \epsilon$, as in \cite{Dai2018}, for $\epsilon \in [0.01,... 0.25]$. The accuracy of label reconstruction was not considered in \cite{Zhou2021}, whereas in \cite{Liu2019c} TPs  only elements with perfectly reproduced labels are counted; this is the reason part of the entries in Table \ref{tab:res_tabular_bars} are empty. We measure accuracy of text label prediction by the ratio of Levenshtein distance $L_{dist}$ \cite{Levenshtein66}, defined as $R_{Lev}(s,t)=(|s|+|t|- L_{dist}(s,t))/(|s|+|t|)$ for strings $s,t$. 
 The exactly recovered labels correspond to $R_{Lev} =1.0$, which we used as a condition for comparison to \cite{Liu2019c} results. 
 We observe that our numerical accuracy for bar charts is lower than that of \cite{Zhou2021} at the more strict $2\%$ error restriction, but is better at $5\%$ test; this may follow from the limited 512x512 resolution of the CenterNet input images. On pie charts (Table \ref{tab:res_tabular_pies} ) our system outperforms \cite{Liu2019c} except at the low accuracy standard $\epsilon=0.25$. Our accuracy saturates at $\epsilon=0.1$, meaning that we don't have errors higher than $10\%$. This behaviour may be more suitable for real-world applications, where inaccurate  numerical predictions can mislead the user systems. 
 
 \begin{table}
  \caption{Accuracy of bar values in extraction of tabular data}
  \label{tab:res_tabular_bars}
  \begin{tabular}{c|cccccc}
    \toprule
     Method & \small{A.L} &\small{A.L}& \small{E.L} &\small{E.L} & \small{E.L} & \small{E.L }\\
     $\epsilon$ & \small{0.02} & \small{0.05} & \small{0.01} & \small{0.05} & \small{0.01} & \small{0.025}\\
    \midrule
     \cite{Zhou2021} & \textbf{67.0\% }  & 71.0\%  & --           &     --    & --     & --     \\
     \cite{Liu2019c}    & --           & --  & 28.4\%         & 32.8 \% & 34.3 \%  & 38.8 \%\\ 
     Ours             & 60.0\%    & \textbf{74.2\%}   & \textbf{31.6}\%      & \textbf{55.8\% }   &  \textbf{58.3\% }   &  \textbf{60.3\% }   \\
    \bottomrule
  \end{tabular}
  \caption*{E.L - exact label prediction, A.L - any labels. }
\end{table}

 \begin{table}
  \caption{Accuracy of pie segment angles in extraction of tabular data from real-world pie charts with exact labels}
    \vskip -2 mm
  \label{tab:res_tabular_pies}
  \begin{tabular}{c|cccc}
    \toprule
    Method           & $\epsilon=0.01$  & $\epsilon=0.05$ & $\epsilon=0.1$ & $\epsilon=0.25$ \\
    \midrule
     \cite{Liu2019c}           & 28.9\%      & 57.8\%     &60.0\%      & \textbf{64.6\%} \\ 
     Ours ($R_{Lev} =1.0$)  & \textbf{44.9}\%      &  \textbf{60.3}\%    & \textbf{61.0}\%        & 61.0\%        \\
    \bottomrule
  \end{tabular}
\end{table}

\subsection{Ablation study}
We explore the impact made by introducing the heatmaps for detection of pie objects, which can be alternatively detected using sectors boxes.
In Table \ref{tab:res_ablation_sectors} we present accuracy of predicted sector angles in pie charts, using only the sector box proposals or using the set of heatmaps detailed in Section \ref{sect:analysis_pies}. The accuracy is substantially improved (up to $10\%$) by using heatmaps-based computation of the graphical elements. 
This table also sheds some light on the distribution of label reconstruction accuracy among the reasonably accurate ($5\%$ rel. error) sectors.
 \begin{table}
  \caption{Ablation study of heatmaps impact for pie charts }
  \label{tab:res_ablation_sectors}
  \begin{tabular}{l|cc}
    \toprule
    Label accuracy & with sector boxes      &   with  heatmaps\\
    \midrule
        $R_{Lev} =1.0$  (exact labels)  & 49.4\%          & 59.7\%         \\
         $R_{Lev} =0.8$   & 58.2\%          & 66.1\%         \\
          $R_{Lev} =0.4$   & 64.2\%          & 71.0\%         \\
          $R_{Lev} =0.0$        (any labels)  & 93.3\%          & 98.4\%    \\
    \bottomrule
  \end{tabular}
\end{table}

\section{Summary and conclusion}
We have presented the CHARTER - a practical system for document charts conversion to tabular data, which can extend the scope of real-world document conversion tools to include the content of the embedded document charts. The proposed use of heatmaps allows more accurate analysis of line, scatter, bar and pie charts within the same model. Moreover, our model can be easily extended to include additional chart types, which is in the scope of our future work.

\clearpage
\bibliographystyle{ACM-Reference-Format}
\bibliography{references.bib}

\end{document}